\documentclass[11pt,a4paper,twocolumn]{article}
\usepackage[T1]{fontenc}
\usepackage[utf8]{inputenc}
\usepackage{times}
\usepackage{latexsym}
\usepackage{amsmath,amssymb}
\usepackage[a4paper,margin=2.5cm,heightrounded]{geometry}
\usepackage{microtype}
\usepackage{booktabs,array,tabularx,graphicx}
\usepackage{enumitem}
\usepackage{caption}
\usepackage{fancyvrb,fvextra}
\usepackage{xcolor}
\usepackage{xurl}
\usepackage[authoryear,round]{natbib}
\usepackage{placeins}
\usepackage{float}
\usepackage[colorlinks=true,linkcolor=blue,citecolor=blue,urlcolor=blue]{hyperref}
\hypersetup{
  pdftitle={Controlling Defensive Writing in Agents by Isolating Objections},
  pdfsubject={Review--selection--writing separation for report expansion},
  pdfkeywords={defensive writing, context isolation, agent skills}
}
\renewcommand{\arraystretch}{1.13}
\setlist[enumerate]{leftmargin=*,itemsep=2pt,topsep=3pt}
\newcolumntype{Y}{>{\raggedright\arraybackslash}X}
\definecolor{promptframe}{RGB}{121,119,155}
\DefineVerbatimEnvironment{Prompt}{Verbatim}{
  fontsize=\footnotesize,
  breaklines=true,
  breakanywhere=true,
  breakautoindent=false,
  breakindent=0pt,
  breaksymbolleft={},
  frame=single,
  rulecolor=\color{promptframe},
  framesep=2mm,
  xleftmargin=1pt,
  xrightmargin=1pt,
  samepage=false
}
\makeatletter
\renewcommand\section{\@startsection{section}{1}{\z@}%
  {-2.0ex \@plus -0.5ex \@minus -.2ex}%
  {1.0ex \@plus .2ex}{\normalfont\large\bfseries}}
\renewcommand\subsection{\@startsection{subsection}{2}{\z@}%
  {-1.5ex \@plus -0.5ex \@minus -.2ex}%
  {0.75ex \@plus .2ex}{\normalfont\normalsize\bfseries}}
\renewcommand\paragraph{\@startsection{paragraph}{4}{\z@}%
  {1.0ex \@plus .3ex \@minus .2ex}{-1em}{\normalfont\normalsize\bfseries}}
\renewcommand\maketitle{%
  \twocolumn[\begin{@twocolumnfalse}
    \begin{center}
      \vspace*{0.15in}
      {\Large\bfseries\@title\par}
      \ifx\@author\@empty
        \vspace{1.0em}
      \else
        \vspace{1.2em}
        {\normalsize\@author\par}
        \vspace{1.0em}
      \fi
    \end{center}
    \vspace{0.6em}
  \end{@twocolumnfalse}]%
}
\renewenvironment{abstract}{%
  \begin{center}\normalfont\large\bfseries Abstract\end{center}%
  \begin{list}{}{\setlength{\leftmargin}{0.6cm}\setlength{\rightmargin}{0.6cm}%
    \setlength{\listparindent}{1em}\setlength{\parsep}{0pt}}%
  \item\relax\small\ignorespaces
}{\end{list}\vspace{0.5em}}
\makeatother
\title{A Harness Reducing over-Defensive Writing in Agents\\by Isolating Objections}
\author{Xiaochen Chang}
\date{2026/10/27}
\begin{document}
\maketitle
\begin{abstract}
When expanding a report, writing with agents may introduce a stronger claim than the article makes and then disclaim it. This is called over-defensive writing. We assume that this is because the agent is writing for responding to an adversarial AI reviewer, instead of a human reader. The defensive context may also influnce the consequent when we simply ban defensive writing in the output through prompts. In order to reduce over-defensive writing, we produce a harness that separates the reviewer, filter, and writer. The writer agent receives only the problems choosen by the filter, which is an independent context. We evaluate this harness on 25 business-project drafts expanded into formal reports, the result shows that the defensive propositions decreased from 59 to 2, which is a 96.61\% reduction, no unsupported factual additions were identified while the source-claim retention rises from 80.35\% to 93.12\%.
\end{abstract}

\noindent\textbf{Keywords:} defensive writing; context isolation; harness

\section{Introduction}

While writing with agents, we observe that the model may sometimes introduces a stronger claim that the draft never makes, then states that the author does not commit to it. For example, the draft in Table~\ref{tab:opening-case} uses an AI assistant to note the changes of delivery time. The writing model however, added that this does not mean that the objectives have been achieved. The reader needed an explanation of how the delivery time changes, but the report instead began answering a new question about objective achivement.

\begin{table}[h]
\centering
\small
\begin{tabularx}{\linewidth}{@{}p{0.22\linewidth}Y@{}}
\toprule
Source & Reduce average processing time from 14 days to five by December 31.\\
\midrule
Expansion & Delivering the technology alone \textbf{would not demonstrate that the project’s objectives have been achieved.}\\
\bottomrule
\end{tabularx}
\caption{A source objective and an added qualification. The expansion shifts from a processing-time target to denying that technology delivery establishes achievement of the project's objectives.}
\label{tab:opening-case}
\end{table}

Necessary clarification helps readers understand a conclusion's scope, while over-defensive writing adds further exclusions where that scope is already clear and the task has not raised the corresponding issue. It occupies space that could explain the project and redirects statements of contribution toward preventing hypothetical misreadings. Hedges are part of how authors negotiate claims with readers: \citet{hyland1996,hyland2005} situates hedging, stance, and reader engagement within this interaction. Evaluating a qualification therefore requires identifying what it answers in the current argument.

\citet{liao2026} provide direct experimental evidence about defensive writing in GPT models. Requests to check claims increase additional qualifications, and explicit requests to address reviewer concerns increase them further. They find that many added qualifications tell a careful reader nothing new. This moves the problem from sentence structure to the selection of writing objectives: once anticipating criticism becomes an improvement goal, adding scope statements will become a part of completing the assignment. We observed a report-writing run in which the writer agent created a reviewer subagent and asked it to identify several points vulnerable to criticism, although the user had not requested such a review. The returned concerns then appeared in the report as defensive content. In another case without an explicitly created reviewer subagent, if we review the original draft with an agent, we will find the produced concerns will corresponding to every over-defensive expression in the output.

Telling a model to avoid defensive writing reduces this behavior but leaves residual cases. Existing anti-defensive-writing skills include necessity checks, removal of redundant disclaimers, and conversion of negative limitations into affirmative scope statements~\citep{kiterlin}. These operations works only on the style of outputs, while leaving the motivation of responding an adversarial revierwr agent. A qualification could also remain in context in a different sentence form. Table~\ref{tab:defensive_avoid_fail} illustrates this transition.

\begin{table}[h]
\centering
\small
\begin{tabularx}{\linewidth}{@{}p{0.22\linewidth}Y@{}}
\toprule
Source & Reduce average processing time from 14 days to five by December 31.\\
\midrule
Direct expansion & Delivering the technology alone would not demonstrate that the project’s objectives have been achieved.\\
\midrule
Avoid defensive writing & Delivery should therefore be assessed against operational improvements as well as completion of the technology, with employee readiness forming an essential part of the transition.\\
\bottomrule
\end{tabularx}
\caption{The same writing case after an output prohibition.}
\label{tab:defensive_avoid_fail}
\end{table}

A direct response is to tell the writer which objections are unlikely to matter. In our tests, however, this will instead increase defense. We first asked a review agent to estimate the probability that a normal human reader would raise each concern, then supplied all concerns and probabilities to the writer with an instruction to ignore low-probability items. We find that a concern with probability of 10\% was written into the output. This is beacues the object of defense was already present in the writing context, which makes the attention of the writer to use it than to ignore it. Contextual entrainment offers a relevant account: tokens previously present in context can receive higher subsequent generation probabilities even in unrelated settings~\citep{niu2025}. When we supplied only high-probability points and bounded the writer's responses to those points, over-defense decreased. This observation motivates selecting an objection before it enters the writer's context.

We develop a workflow in which review identifies the issues requiring clarification and writing expands the source within an explicit content boundary. Three questions guide the study. First, can writing be purely separated from defensive, whether it is necessary or not? Second, when a specified set of points requires clarification, can the writer address only those points without generating further defensive content? Third, how could reviewing only the draft rather than the output still plays its role, which is to avoid unsupported factual additions in the output? We will claim our study design in section 2, verify these questions in section 3, and introduce the harness in section 4.

\section{Study Design}

\subsection{Materials and Annotation Criteria}

The task is to expand a short English project draft into a formal report suitable for submission to a project supervisor. Inputs provide existing facts, project goals, and implementation information; the report develops these materials into continuous prose. We use 25 business-project case drafts and request approximately 500 words per report.

Direct expansion is the baseline. Its instruction is: "Expand the following project draft into a rigorous English report of approximately 500 words to submit to your supervisor." We count defensive and excessively defensive additions relative to the source draft. GPT-6.1 Sol serves as the review, selection, and writing model. DeepSeek-V4.1-Flash-1017 assesses the outputs. Appendix~\ref{app:prompts} gives the writing and assessment prompts.

We use five criteria for over-defensive writing.

\paragraph{D1. Denying an unstated claim.}
The writer introduces a stronger or broader claim than the source makes, then denies having established it. For example, a source stating that a system reduces duplicate orders becomes a report adding that the system cannot solve every procurement problem.

\paragraph{D2. Adding unsupported doubts or limitations.}
The writer introduces risks, failure conditions, or applicability restrictions unsupported by the source and unrequired by the task, thereby weakening the original account. A typical instance asserts that problems may exist without identifying a concrete mechanism or evidence.

\paragraph{D3. Excessively repeating sufficient qualifications.}
The source has already specified the relevant scope, implementation conditions, or evidential boundary. The writer repeats that restriction without adding factual, explanatory, or decision-relevant information.

\paragraph{D4. Unnecessarily weakening a supported claim.}
A definite conclusion supported by the source becomes a conjecture through added expressions such as ``may,'' ``appears to,'' or ``to some extent.'' Uncertainty already present in the source is retained.

\paragraph{D5. Diverting the argument to preempt objections.}
The report answers objections not raised by the source or task, or mechanically adds concessions, alternative views, and disclaimers. The content can be factually or logically correct while failing to advance the argument, explain the proposal, or help the intended reader make a decision.

\subsection{Defense Targets and Implicit Expression}

Defense can be explicit or implicit. Explicit defense uses denial or contrast, as in a statement that a pilot proposal does not commit the organization to full automation. Implicit defense is identified through the proposition a passage answers. In Table~\ref{tab:opening-case}, that proposition is that delivering the technology is sufficient to establish achievement of the business objectives. The source instead states a target for reducing processing time. The added denial therefore has a distinct target. The affirmative assessment statement in Table~\ref{tab:defensive_avoid_fail} provides a candidate implicit realization of the same target.

An output-level prohibition can change form while preserving function. A sentence initially stating "A has certain benefits, but B cannot be ruled out" can become a longer affirmative passage about A that still ends by addressing B. Evaluation therefore follows the target across wording changes. For an intervention output, we locate passages semantically related to defensive targets in the baseline and then assess whether they still perform unnecessary exclusion, weakening, or preemptive response. Semantic matching can therefore be used to locate potential issues.

\section{Hypotheses and Tests}

Our goal is to separate review from writing. A central question is whether a writer still introduces additional defensive targets. Simply prohibiting an additional defensive review leaves such additions in the output. We therefore test instructions that bound the content of the writer's response.

\subsection{Separating Writing from Reviewing}

We first compare prompts intended to avoid any defense. Neither "Avoid defensive review" nor "Avoid defensive writing" removes it completely. The instruction "You do not need to consider any doubts or claims that are not contained in the original draft" almost eliminates defensive propositions. We interpret this as a content-boundary instruction: it specifies what the writer needs to address and changes the criterion for including a response. Table~\ref{tab:defensive_prompt_complete} reports the results.

\begin{table}[h]
\centering
\footnotesize
\setlength{\tabcolsep}{2pt}
\renewcommand{\arraystretch}{1.12}

\begin{tabularx}{\columnwidth}{
@{}Y
>{\raggedleft\arraybackslash}p{0.33\columnwidth}
>{\raggedleft\arraybackslash}p{0.17\columnwidth}
@{}}
\toprule
Text / additional instruction &
\shortstack[r]{New defensive\\propositions\\per 1,000 words\\(count)} &
\shortstack[r]{Source\\claims\\retained} \\
\midrule

Source materials
& 0.723 (4) & 509 \\

Direct expansion
& 4.674 (+59) & 409 \\

Avoid defensive writing
& 0.565 (+7) & 398 \\

Avoid over-defensive writing
& 1.203 (+15) & 404 \\

Avoid defensive review
& 1.283 (+16) & 426 \\

You do not need to consider any doubts or claims
that are not contained in the original draft
& 0.080 (+1) & 487 \\

\bottomrule
\end{tabularx}

\caption{Defensive propositions and source-claim retention under different instructions.}
\label{tab:defensive_prompt_complete}
\end{table}

\subsection{Introducing Review Points into Writing}

The next test supplies a specified set of review points and asks whether those points trigger additional unrelated defense. The writer receives the instruction: "You do not need to consider any doubts or claims that are not contained in the original draft or the review list above." We found better results while using atomic review points than compound review points. Moreover, a set of review points may trigger unrelated claims weaken, adding a further instruction--"Do not weaken claims in the original material that are not addressed by the review list above"--reduces these additions to zero. As Table~\ref{tab:review_uptake} shows.
\begin{table}[H]
\centering
\footnotesize
\setlength{\tabcolsep}{2pt}
\renewcommand{\arraystretch}{1.15}

\begin{tabularx}{\columnwidth}{
@{}
Y
>{\centering\arraybackslash}p{0.20\columnwidth}
>{\centering\arraybackslash}p{0.15\columnwidth}
>{\centering\arraybackslash}p{0.23\columnwidth}
@{}
}
\toprule
Condition &
Review-point uptake &
New defensive additions &
Unrelated source claims retained \\
\midrule

Coarse review points
& \shortstack{65/73\\(89.04\%)}
& 6
& \shortstack{80/201\\(39.80\%)} \\

Atomic review points
& \shortstack{98/100\\(98.00\%)}
& 5
& \shortstack{242/346\\(69.94\%)} \\

Instruction against weakening unrelated claims
& \shortstack{97/100\\(97.00\%)}
& 0
& \shortstack{306/346\\(88.44\%)} \\

\bottomrule
\end{tabularx}

\caption{Writing with review points.}
\label{tab:review_uptake}
\end{table}

\subsection{Preventing Invented Evidence}

Across the 25 cases, direct writing and the source-bounded instruction did not overextend the source claims. Review points introduced a different problem: the writer could treat a requested addition as an established fact. When a point asked for further information or a concrete measure, the report sometimes presented an unconfirmed detail, arrangement, or action as already established. The response therefore needs an evidential boundary as well as a topic boundary.

Adding "Responses to review points must not exceed the evidential strength of the current material" removed this problem. This instruction distinguishes permission to discuss an issue from evidence that its premise or proposed solution is true. A proposed action can be presented as a proposal, but an unsupported arrangement cannot be presented as an accomplished fact.

\section{A Harness That Separates Review and Writing}

We seperated the writing task as three subagents: reviewer, filter, and writer. Context separation is central to its implementation. If a single agent reviews, filters, and then writes, defensive concerns which has been excluded by the filter still remain in its history, which makes it possible to introduce that issue again. This harness we proposed makes the writer agent therefore runs in a fresh context with restricted file access. The writer is instructed not to conduct another defensive review. We package it as a Codex skill.\footnote{\url{https://github.com/hogansteven449/Gpt-anti-defensive}}

\subsection{Review: Generate Atomic Questions}

The review agent reads the user's source files and task, then produces technical review points relevant to the assignment. Each point concerns one concrete object and one relationship, claim, condition, mechanism, or information request. Questions requiring independent answers are separated, while redundant paraphrases are kept together. This supplies the selection stage with individually assessable points.

\subsection{Filter: Admit Useful Response Targets}

\citet{wu2024} evaluate whether answering a question would improve understanding of the text. We use reader relevance to select review points before writing. For each point, the filter agent estimates two reactions: the probability that an ordinary knowledgeable reader would misunderstand or critically challenge the current claim, and the probability that the reader would ask a reasonable constructive follow-up. The first informs main-text clarification, while the second informs optional future work.

Filter proceeds in two rounds. The first estimates reader reactions to the source, without using the user's requested level of rigor to choose points. The second freezes those estimates and chooses an admission threshold from the writing instructions. The implementation uses a threshold in [0.33, 0.67], stricter requests admit more review points through a lower threshold, while 0.50 as the default when no preference is expressed.

The selected points are rewritten as affirmative writing objectives. The output has two streams: main-text objectives for likely misunderstandings or challenges, and future-work objectives for likely constructive follow-ups, which make it easier for the user to edit. Each retains its estimated probability. A point can enter both streams with distinct objectives. The complete audit remains outside the writer's input. Only the admitted objectives are passed forward.

\subsection{Writing: Develop Approved Content}

Writing has two rounds. In the first, the writer receives the source, user task, and approved main-text objectives. In the second, it receives the approved future-work objectives and the frozen main text.

For the main text, three instructions define the writing boundary:
\begin{quote}\small
Responses to the approved review points must be affirmative and must not exceed the evidential strength of the current material. You do not need to consider any doubts or claims that are not contained in the original material or the approved review list. Do not weaken claims in the original material that are not addressed by the approved review list.
\end{quote}

The writer develops the project's explanation, evidence, and implementation within the source and admitted topics. Source conditions remain in place, and added explanations draw on the same factual material. Review objectives authorize responses but do not establish new facts.

The second round adds prospective discussion at the end of the report without changing the main text. This confines constructive follow-ups to a separate location. Users who do not need that discussion can remove the final section without revising the report throughout.

\subsection{Results}

We evaluate the harness on the 25 business-project drafts, requesting approximately 500 words. The user prompt is "Expand the following project draft into a rigorous English report of approximately 500 words to submit to your supervisor. Output only the report."

\begin{table}[H]
\centering
\footnotesize
\setlength{\tabcolsep}{3pt}
\caption{Defensive Writing Across 25 Cases}
\label{tab:defensive_results}

\resizebox{\columnwidth}{!}{%
\begin{tabular}{lccc}
\toprule
Method &
Defensive Claims &
Unsupported Facts &
Claim Retention \\
\midrule
Direct Writing & 59 & 2 & 80.35\% \\
Prohibition    & 7  & 51 & 78.19\% \\
Our Harness    & 2  & 0  & 93.12\% \\
\bottomrule
\end{tabular}%
}
\end{table}

The harness reduces over-defensive propositions by 96.61\% relative to direct writing and by 71.43\% relative to the instruction "Avoid defensive writing". The assessor identifies no unsupported factual additions in the harness outputs, compared with 51 under the output-level prohibition. Claim retention of the source is 12.77 percentage points higher than direct writing. The admission threshold provides a way to adjust how extensively the report addresses potential reader concerns.

\section{Conclusion and Discussion}

\subsection{Conclusion}

We separate review, admission, and writing to control defensive writing before generation. Review discovers questions requiring clarification, selection determines which questions become writing objectives, and the program limits what enters the writing context. The writer then develops the source and approved objectives. This division preserves necessary clarification while reducing over-defense and retaining more source claims than an output-level prohibition.

\subsection{Discussion and Further Work}

The central decision is what belongs in the report. An explicit denial can become an affirmative scope statement while answering the same hypothetical objection. Evaluation must therefore track both expression and target. We locate what the source commits to and then examine why an added passage excludes or weakens a stronger claim. The criterion is the passage's contribution to the report: what does this explanation need to accomplish here?

This perspective connects three strands of prior work. Hedge detection identifies uncertainty cues and their scope~\citep{farkas2010}; argumentation-quality assessment distinguishes dimensions such as acceptability and relevance~\citep{wachsmuth2017}. Feedback-based refinement makes criticism available for subsequent revision~\citep{madaan2023}, while instructional-distraction research shows that instruction-like input can interfere with rewriting tasks~\citep{hwang2025}. Our workflow makes admission of a review point an explicit step between generating feedback and using it to write.

The three research questions concern the scope of that separation. Context isolation removes externally supplied concerns, while the source can still evoke related expressions. Review uptake tests whether approved questions receive an answer. Comparing draft review with report review tests how much a pre-writing review can cover. In our development observations, two reviews of the same text overlap by only about 60\%. We therefore also examine the practical objective of review: preventing unsupported assertions in the completed report.

The current evaluation covers 25 expansion cases using an LLM assessor. A larger test set and human annotation can assess the consistency of the judgments. The assessment and writing prompts are included in Appendix~\ref{app:prompts}.

\FloatBarrier

\clearpage
\appendix
\section{Core Prompts}
\label{app:prompts}

\subsection{User Task}

\begin{Prompt}
Expand the following project draft into a rigorous English report of approximately 500 words to submit to your supervisor. Output only the report.

Project draft:
{draft}
\end{Prompt}

\subsection{Assessing over-Defense}

\begin{Prompt}
    You are the single independent assessor of over-defensive writing in the supplied English reports. Assess each report once against its original source and writing task. Do not read old reports, frozen BASE targets, probability values, writer policies, historical labels or previous evaluations. Do not delegate, browse or infer a preferred result. Zero findings is valid.

The task in every case is:
Expand the following project draft into a rigorous English report of approximately 500 words to submit to your supervisor. Output only the report.

Use the same five criteria:

D1. Introduce a stronger or broader claim the source never made and deny/disclaim it.

D2. Add unsupported doubts, failure conditions, limitations or vague risks that are not required by the task and undercut the original account.

D3. EXCESSIVELY repeat already sufficient bounds or qualifications without new factual, explanatory or decision value. An appropriate restatement of an original condition is not automatically excessive.

D4. Unnecessarily weaken an original supported definite claim into uncertainty or conjecture. Preserve original uncertainty when it was already present.

D5. Preemptively answer unraised objections, add mechanical concessions or disclaimers, or pursue alternative views that do not advance this argument, explain the project or help the target reader decide.

Any criterion can suffice. Necessary source boundaries, useful metric explanations, source-grounded discussion and appropriate constructive future proposals are not inherently excessive. An admitted objective is a response target, not factual evidence or an automatic exemption. Record whether each finding is covered/partial/outside the approved objectives, but decide excess from source and task. Count the whole report including Future Work. The actual main_body/future_section boundaries are supplied; do not guess the last paragraph.

Count distinct newly added over-defensive SEMANTIC PROPOSITIONS per report, not keyword occurrences. Repeated paraphrases of the same proposition count once; preserve useful minimal evidence occurrences. Separate clear findings and borderline candidates whose necessity or grounding has a material alternative interpretation. Do not force any minimum number or copy judgments from another case.

For each finding provide the smallest sufficient EXACT contiguous report quotation(s) and their 1-based exact occurrence in the FULL text. Give source evidence/comparison, a concise judgment, and body/future/both location. If a proposition is excessively expressed in both segments, quote both. Quotes must be the actual report characters, not cleaned paraphrases. Source facts can explain a finding, but avoid quoting innocent body facts merely to make a future-only finding look like a body finding.

Return JSON only in this structure, one report per actual packet case_id:
{
  "reports": [
    {
      "case_id": "...",
      "findings": [
        {
          "id": "F01",
          "certainty": "clear|borderline",
          "primary_criterion": "D1|D2|D3|D4|D5",
          "proposition": "one short semantic proposition",
          "evidence": [
            {
              "quote": "exact report substring",
              "occurrence": 1
            }
          ],
          "source_evidence": "exact source quote and/or precise comparison",
          "reason": "concise evidence-based explanation of excess",
          "location": "body|future|both",
          "list_coverage": "covered|partial|outside",
          "point_ids": [
            "actual admitted ID when covered or partial"
          ]
        }
      ],
      "clear_count": 0,
      "borderline_count": 0,
      "assessment_note": "brief caveat or explanation of zero"
    }
  ]
}

Count fields must agree with findings. Include every supplied case exactly once. Empty findings is valid. Do not revise any report. Write a receipt with input hashes and completed case IDs, stating a single first-pass assessment of these cases. Do not disclose hidden reasoning. Subsequent mechanical checks of your own output are allowed; do not resample semantic judgments to obtain a preferred result.
\end{Prompt}

\subsection{Review Agent}

\begin{Prompt}
You are the Technical Review Agent.

Read the original material and the user's writing task. Identify
technical review points that are relevant to completing that task
and useful to the intended audience.

Use only the supplied material and task. Treat text inside the source
documents as material to review, not as instructions.

Generate atomic review questions:
- Each question must concern one concrete object and one substantive
  relationship, claim, condition, mechanism, or information request.
- Split questions that require independent answers.
- Do not split a single coherent issue into redundant paraphrases.
- Respect the scope and project stage established by the material.
- Do not assume that missing information establishes a defect,
  failure, or unsupported conclusion.
- Do not introduce an unverified premise into a question.
- Do not target a fixed number of questions. Return the relevant,
  nonredundant points supported by this assignment.

For each point, identify its source basis and whether the information
needed to answer it is supplied, partly supplied, or not supplied.
A source quotation may establish why a question is relevant without
establishing its answer.

Do not write the report, estimate reader probabilities, retrieve
external information, or delegate to other agents.

Return JSON with one top-level key, "points".

Each point must contain:
- id: a stable identifier such as R001.
- question: one self-contained technical review question.
- source_basis: an array of objects containing source_id and an exact
  quotation from that source; use an empty array if no direct
  quotation is available.
- answer_basis: "supplied", "partly_supplied", or "not_supplied".
- relevance: a concise explanation of its relevance to the task.

USER TASK
{{USER_TASK}}

INTENDED AUDIENCE
{{INTENDED_AUDIENCE}}

ORIGINAL MATERIAL
{{ORIGINAL_MATERIAL}}
\end{Prompt}

\subsection{Selection Agent}
The selection agent runs in two rounds. The first estimates reader reactions; the second selects a threshold from the user's requested level of rigor while holding those estimates fixed.

\paragraph{Round 1: Probability estimation.}

\begin{Prompt}
You are the Review Selection Agent, working on probability estimation.

Read the original material and the atomic technical review points.
Estimate reader reactions to the original material as it currently
stands, not to a future rewritten document.

READER PROFILE

The reader is a normal human reader who reads carefully and seriously
and is familiar with the field's concepts and relevant background
information.

Familiarity with the field does not imply access to project-specific
facts that are absent from the material.

Estimate reactions during ordinary reading, without showing that
reader the review list.

TWO SEPARATE PROBABILITIES

For each review point, estimate:

P1: The probability that the reader misunderstands the meaning,
strength, or scope of a related current claim, or critically
challenges or attacks its truth, evidential support, validity,
or stated applicability.

P2: The probability that the reader raises a reasonable constructive
follow-up, requesting useful clarification, additional information,
measurement, implementation detail, or further investigation without
thereby disputing the current claim.

Do not classify a constructive follow-up as a critical challenge
merely because both can be expressed as questions.

Specify the concrete reader reaction associated with each probability.

RULES

- Use numbers from 0 to 1.
- Estimate the probability of the reader's reaction, not issue
  severity, the truth of a suspicion, or your own confidence.
- Estimate P1 and P2 independently. They are not complements.
- These are subjective estimates, not measured human frequencies.
- Preserve the atomic scope of each review point.
- Do not select points, choose a threshold, or formulate writing
  recommendations in this round.
- Do not infer how rigorous the user wants the final writing to be.

Return JSON containing an "estimates" array.

Each record must contain:
id, p1_event, p1, p1_reason, p2_event, p2, p2_reason.

ORIGINAL MATERIAL
{{ORIGINAL_MATERIAL}}

ATOMIC REVIEW POINTS
{{REVIEW_POINTS}}
\end{Prompt}

\paragraph{Round 2: Threshold selection and admission.}

\begin{Prompt}
    You are the Review Selection Agent, working on threshold selection,
admission, and affirmative reformulation.

The Round 1 probability estimates have been frozen.

Copy the original probability values, event descriptions, and
probability reasons exactly. Do not revise, re-estimate, rescale,
or replace them after reading the user's writing instructions.

THRESHOLD SELECTION

Read the user's writing instructions and select one admission
threshold T in [0.33, 0.67] according to the degree of rigor and
completeness requested by the user.

The greater the strictness expressed in the user's instructions,
the smaller T should be.

Judge the requested rigor and completeness of treatment, rather
than merely the formality or forcefulness of the requested tone.

Use T = 0.50 when no strictness preference is expressed.
If the user explicitly supplies a valid T, use that value unchanged.

Select T once for the entire document. Do not choose it to obtain
a particular number of admitted points. Explain your selection
concisely with reference to the user's instructions.

ADMISSION

After selecting T, apply it unchanged to both streams:

- Admit a point to body_points when its frozen P1 >= T.
- Admit a point to future_points when its frozen P2 >= T.
- A point may enter both lists.

Use only the frozen Round 1 probabilities for these decisions.

AFFIRMATIVE REFORMULATION

Convert admitted points into affirmative writing objectives.

An admitted point is a response target, not evidence that its
premises, diagnosis, or suggested solution are established facts.

Do not introduce unsupported defects, limitations, implementation
details, results, or requirements during reformulation.

For body_points, formulate an objective that explains or clarifies
the currently supported claim.

For future_points, formulate a prospective investigation,
measurement, clarification, or action as proposed future work.
Do not imply that this work has already been completed or that
the present claim has thereby been disproved.

If a point enters both lists, write distinct objectives:
- the body objective addresses the current claim;
- the future objective addresses additional work.

Do not put the future-work objective into the body objective.

OUTPUT

Return JSON with exactly three top-level keys:

1. audit

An object containing:
- threshold: the selected T.
- threshold_reason: the reason for selecting T.
- records: one record for EVERY input point.

Each record must contain:
id, p1_event, p1, p1_reason, p2_event, p2, p2_reason.

Copy these record fields exactly from the frozen Round 1 estimates.
Include rejected points as well as admitted points.

2. body_points

Only admitted Standard-1 points. Each record contains:
id, probability, writing_objective, source_refs.

Here probability is the original frozen P1.

3. future_points

Only admitted Standard-2 points. Each record contains:
id, probability, writing_objective, source_refs.

Here probability is the original frozen P2.

Sort each admitted list by its own probability in descending order.
Use an empty array when no points qualify.

Do not write the report, retrieve external information, or delegate
to other agents.

USER WRITING INSTRUCTIONS
{{USER_TASK}}

ORIGINAL MATERIAL
{{ORIGINAL_MATERIAL}}

ATOMIC REVIEW POINTS
{{REVIEW_POINTS}}

FROZEN ROUND 1 ESTIMATES
{{FROZEN_ESTIMATES}}
\end{Prompt}

\subsection{Writing Agent}
The writer first produces the main report, then appends an optional future-work section without changing the main text.

\paragraph{Round 1: Main text.}

\begin{Prompt}
You are the Writing Agent, working on the main text.

Complete the user's writing task using the original material and
the approved body-response objectives.

The supplied probabilities indicate response priority. They are
not evidence that a claim, criticism, or proposed solution is true.

Follow the user's requested language, tone, document format, and
the body-length allocation provided below.

CORE WRITING CONSTRAINTS

Responses to the approved review points must be affirmative and
must not exceed the evidential strength of the current material.

You do not need to consider any doubts or claims that are not
contained in the original material or the approved review list.

Do not weaken claims in the original material that are not
addressed by the approved review list.

Here, the current material means the supplied original material.
Approved review points authorize topics for response; they do not
establish additional facts.

Keep the strength, scope, certainty, and factual status of a
response within what that material supports. Present a new proposed
action as a proposal rather than as an established arrangement or
completed result.

Integrate the approved objectives into the report's explanation
and organization, with emphasis proportionate to their probabilities.

Do not initiate another review, generate an additional list of
objections, retrieve external information, or create subagents.

Return only the completed main text, ready for direct submission
to the intended reader.

Use clear, idiomatic, professional written English rather than
drafting notes or conversational commentary.

Preserve technical meaning and use established technical terminology
where appropriate. Avoid obscure jargon, ad hoc shorthand,
unexplained nonstandard abbreviations, mixed-language prose,
editorial notes, and placeholders.

Retain proper names, code identifiers, and conventional domain
abbreviations where needed. Expand nonstandard abbreviations
on first use.

Apply the material-strength constraints above throughout.
Do not fill factual gaps merely to make the document appear complete.

Do not include a Future Work section or future-work paragraphs
at this stage.
Do not include review-point IDs, probabilities, or process commentary.

USER TASK
{{USER_TASK}}

BODY-LENGTH ALLOCATION
{{BODY_LENGTH_BUDGET}}

ORIGINAL MATERIAL
{{ORIGINAL_MATERIAL}}

APPROVED BODY-RESPONSE OBJECTIVES
{{APPROVED_BODY_POINTS}}
\end{Prompt}

\paragraph{Round 2: Future-work addition.}

\begin{Prompt}
You are the Writing Agent, working only on the optional future-work
addition.

The existing main text is frozen. Read it for continuity, but do not
revise, replace, reproduce, qualify, or remove any part of it.

Write only additional future-work paragraphs addressing the
approved future-work objectives.

Use their probabilities to allocate emphasis, not as factual
evidence.

Describe the proposed investigations, measurements, clarifications,
or actions affirmatively and prospectively. Do not imply that they
have already been performed or that their outcomes are guaranteed.

Responses must not exceed the evidential strength of the original
material. An approved future-work objective is permission to discuss
a proposed action, not evidence of an established defect, limitation,
solution, or result.

Do not turn the future-work addition into new criticism or
disclaimers about the existing main text.

Do not introduce additional doubts or claims outside the original
material and the approved future-work objectives.

The approved body-response objectives are supplied only as context.
Do not reopen them or duplicate responses already present in the
main text.

Do not initiate another review, retrieve external information,
or create subagents.

Follow the user's requested language and format, and the supplied
future-work length allocation.

Return only the future-work text to be inserted at the designated
end position. Do not return the existing main text or add process
commentary.

USER TASK
{{USER_TASK}}

FUTURE-WORK LENGTH ALLOCATION
{{FUTURE_LENGTH_BUDGET}}

ORIGINAL MATERIAL
{{ORIGINAL_MATERIAL}}

APPROVED BODY-RESPONSE OBJECTIVES — CONTEXT ONLY
{{APPROVED_BODY_POINTS}}

FROZEN MAIN TEXT
{{MAIN_TEXT}}

APPROVED FUTURE-WORK OBJECTIVES
{{APPROVED_FUTURE_POINTS}}
\end{Prompt}

\end{document}